# A Novel Hybrid GWO with WOA for Global Numerical Optimization and Solving Pressure Vessel Design

Hardi M. Mohammed[1,2], Tarik A. Rashid[3]
[1] Applied Computer Department, College of Medicals and Applied Sciences, Charmo University, Sulaimani, Chamchamal, KRG, Iraq.
[2] Technical College of Informatics, Sulaimani Polytechnic University, Sulaimani, KRG, Iraq.
hardi.mohammed@charmouniversity.org
[3] Computer Science and Engineering, School of Sciecne and Engieering, University of Kurdistan Hewler, Hewler, KRG, Iraq.

## Abstract

A recent metaheuristic algorithm, such as Whale Optimization Algorithm (WOA), was proposed. The idea of proposing this algorithm belongs to the hunting behavior of the humpback whale. However, WOA suffers from poor performance in the exploitation phase and stagnates in the local best solution. Grey Wolf Optimization (GWO) is a very competitive algorithm comparing to other common metaheuristic algorithms as it has a super performance in the exploitation phase while it is tested on unimodal benchmark functions. Therefore, the aim of this paper is to hybridize GWO with WOA to overcome the problems. GWO can perform well in exploiting optimal solutions. In this paper, a hybridized WOA with GWO which is called WOAGWO is presented. The proposed hybridized model consists of two steps. Firstly, the hunting mechanism of GWO is embedded into the WOA exploitation phase with a new condition which is related to GWO. Secondly, a new technique is added to the exploration phase to improve the solution after each iteration. Experimentations are tested on three different standard test functions which are called benchmark functions: 23 common functions, 25 CEC2005 functions and 10 CEC2019 functions. The proposed WOAGWO is also evaluated against original WOA, GWO and three other commonly used algorithms. Results show that WOAGWO outperforms other algorithms depending on the Wilcoxon rank-sum test. Finally, WOAGWO is likewise applied to solve an engineering problem such as pressure vessel design. Then the results prove that WOAGWO achieves optimum solution which is better than WOA and Fitness Dependent Optimizer (FDO).

## Acknowledgments

The authors wish to thank the Sulaimani Polytechnic University and the University of Kurdistan Hewler (UKH).

## 1. Introduction

Optimization is the process to discover an optimum solution in a feasible time. This area has been very dynamic since proposing a Genetic Algorithm (GA) and Differential Evolution (DE). Therefore, the number of optimization problems are increasing and becoming more complex. Consequently, these problems require better optimization methods in order to be solved [1]. There might be several efficient algorithms that can be used to solve a specific problem. However, we cannot consider naming one of them as the best before evaluating it against the others on the problem. As a result, optimization algorithms can be used to solve different problems effectively [2]. There are two types of optimization algorithms: randomized and deterministic. The process of executing deterministic requires at most one direction towards the solution, otherwise; it is terminated. However, the randomized or stochastic technique executes randomly and violates the deterministic constraints [3], [4]. Overall, stochastic is classified as heuristic and metaheuristic. Nature-inspired metaheuristic algorithms can solve real-world problems and standard mathematical functions efficiently in their exploration and exploitation phases. However, balancing between these two phases is a crucial problem in which metaheuristic optimizations are suffered from [5].

NP-hard problems have been solved by most recent metaheuristic algorithms such as job scheduling problem [6], task assignment problems [7], quadratic assignment [8], travel salesman person [9], vehicle routing problem [10], home health care scheduling problem[11] and frequency assignment problem [12]. The most common algorithms are namely Evolutionary Algorithm (specifically GA) [13], Particle Swarm Optimizer (PSO)[14], Artificial Bee Colony (ABC) [15] Whale Optimization Algorithm (WOA) [16] and Grey Wolf Optimizer (GWO) [17].

Mirjalili and Lewis proposed WOA in [16]. This metaheuristic algorithm is motivated by the humpback whale hunting mechanism. This algorithm presented significant results against other metaheuristic algorithms as WOA has random numbers to balance between its two phases. It has better exploration capability by using an updating mechanism. It also uses a random search mechanism in order to change the position for finding optimum solutions. Both exploration and exploitation execute independently so that WOA can avoid local optima and obtain better convergence speed. Despite this mechanism in WOA, other common metaheuristic algorithms do not have specific operators to split the exploration and exploitation, so they fallouts into local optima [16]. WOA has better performance against PSO, Gravitational Search Algorithm (GSA), DE and Feedforward Error Propagation (FEP) [16]. According to [18], WOA performs well in terms of convergence time and balancing between exploration and exploitation.

Despite having efficient performance against common algorithms, WOA has some drawbacks. For example, using a randomization mechanism in WOA for complex problems increases computational time [19]. Convergence and speed are not efficient in both of the phases because they depend on a single parameter which is $a$ [20]. WOA results in poor performance in jumping out from local solutions as the encircling mechanism is used in the search space [21]. Furthermore, not improving the best solution in a better way is another issue that is related to the encircling mechanism [22], and also the WOA exploitation phase requires improvement in order to obtain better solutions. In addition, controlling parameters is crucial in order to improve the performance of the algorithm. It is worth mentioning that the parameter setting has three categories according to their strategies. These are determinist, adaptive and self-adaptive control parameters. Each one of them has an effect on the performance of the algorithm setting. Consequently, these strategies are useful to be involved to improve WOA [23].

As a result, WOA has been hybridized with various algorithms. For example, WOA is integrated with the Local Search (LS) strategy in order to tackle the ordering form of the flow shop scheduling problem. A swap mutation operator is also used to diversify the population to improve performance. Furthermore, WOA could escape from local optima by adding reverse-inserted operation. Therefore, the proposed Hybrid Whale Algorithm (HWA) improved the performance and the solution quality of WOA due to using LS [24].

WOA is also hybridized with Colliding Bodies Optimization (CBO) due to improving the solution quality and convergence rate. In WOA-CBO, whales are divided into two groups which are explorer and imitator.

This division derived from the original CBO. Explorer is those agents who are in the range of lower half whale. However, the upper half is called imitator. Explorer whale changes its position according to the best solution while the imitator updates its position depending on the other half of the whale which is lower [25].

Brain Storm optimization (BS) is hybridized with WOA to tackle the difficulty of stagnation in local optima which WOA has it. In BSWOA, the BS update function is added inside the WOA to update the position of whale based on a coefficient and search area [26].

PSO is an efficient algorithm in the exploitation phase. Thus, it is embedded inside WOA for the exploitation phase while WOA only works in the exploration phase. Therefore, the hybridized algorithm improved and produced better results comparing to WOA and PSO [27].

Because of having problems regarding local optima, the BAT algorithm is used with WOA for the exploration phase. The result of WOA-BAT showed that WOA-BAT improved well comparing to WOA and BAT algorithms [18]. The more detail of WOA modification and hybridization have been described in [18].

Grey wolf optimization was proposed in 2014 in [17]. It is a metaheuristic technique which is inspired by the grey wolves' behavior. This algorithm shows a competitive result against other metaheuristic algorithms. For instance, PSO, DE, GSA and FEP [28]. It is very competitive in the exploitation phase compared to others while it has merit results in the exploration phase. It is also presented a better performance in half of the 29 functions due to avoiding local optima [17].

Despite the fact of having better performance, GWO has issues relating to the balance between exploration and exploitation [29]. It also has a drawback because of having the inability to solve nonlinear equation systems and unconstrained optimization problems [30]. It has an efficient updating mechanism. Though, this mechanism can be improved and enhanced [31]. Initializing grey wolves' population is randomized in order to diversify the population. Still, this practice had a drawback and it was solved in [32].

WOA has the following issues [21], [28] which are the main motivations of hybridizing GWO with WOA in this paper:

1) WOA suffers from avoiding local optima as it uses encircling search mechanism.
2) Improving solution in WOA after each iteration is not sufficient.
3) WOA has low performance in the exploitation phase.

The above problems, which WOA has, motivated authors for proposing the hybridized algorithm. Consequently, authors have decided to choose a hybridized of GWO and WOA to produce better performance in the exploitation phase by GWO, especially when it is evaluated by unimodal benchmark functions. GWO also has a greater capability of exploitation by using multimodal benchmark functions. Thus, this paper aims to propose a hybridized approach to overcome the WOA problems by using two effective ways: the first step is saving the best solution for each iteration, and the second step is comparing each new solution against the best solution in the exploration phase. If the result is better than the best solution, the positions of the agents will get changed, otherwise; they are staying in the old positions. Adding the GWO hunting mechanism in the exploitation phase is the second method in order to enhance the performance of WOA.

The proposed WOAGWO is differentiated from the above hybridizations of WOA as WOA has not been hybridized with GWO. This hybridization combines two techniques (WOA with GWO) and adds a condition to update positions inside the exploitation phase. The proposed algorithm is also distinguished from WOA and GWO as a new update method is added to the exploration phase of WOA. GWO hunting mechanism for updating the position of the whale is also added to the exploitation phase. As a result, WOAGWO is a new suggested hybridization which enhances the performance of WOA.

.

The structure of our paper is organized as WOA with its mechanism is presented in Section 2 and then WOA modifications and hybridizations are explained. After that, GWO modification and hybridization are described in Section 3. Our proposed approach WOAGWO is described in detail in Section 4. In Section 5, WOAGWO is evaluated against 23 common benchmark test functions [16], 25 benchmark test functions from CEC2005 and 10 benchmark functions in CEC2019. Next, statistical results are presented. Furthermore, it is evaluated against other common algorithms, for example, DE, ABC, BSO and WOA. Then, WOAGWO is presented to solve an engineering problem namely: Pressure Vessel Design Problem. Finally, the conclusion with future works is presented.

## 2. WOA

A meta-heuristic algorithm such as WOA is derived from whale behavior. Mirjalili and Lewis first developed this algorithm [16]. It can be said that the school of small fish that are swimming close to the surface of the water is the target to be hunted by a humpback whale. The whale is creating bubbles by shrinking its circle so these circles can be called 9 shaped paths. This algorithm is divided into two phases. The exploration is the first phase which includes the random strategy for searching the prey. Encircling prey can be done in the second phase with the spiral bubble-net attack. This phase is also called the exploitation phase. The following subsections represent details of each phase of WOA[33].

### 2.1 Encircling Prey and Bubble-net Attacking Mechanism

In order to begin the hunt, the whale must first locate the prey. The whale's position is not optimized. Therefore, the whale required to change its position to encircle the prey by using Equation (1) and (2).

$$\vec{X}(i+1) = \vec{X}^*(i) - \vec{A} \cdot \vec{D} \tag{1}$$

$$\vec{D} = \mid \vec{C} \cdot \vec{X}^*(i) - \vec{X}(i) \mid \tag{2}$$

where $\vec{X}^*(i)$ represents the best position of the whale which is found so far at iteration $i$. The current position of the whale is indicated by $\vec{X}(i+1)$, the distance between whale and prey is represented by $\vec{D}$ vector with an absolute value. Coefficient vectors like $C$ and $A$ are calculated respectively:

$$\vec{A} = 2 \cdot \vec{a} \cdot \vec{r} + \vec{a} \tag{3}$$

$$\vec{C} = 2 \cdot \vec{r} \tag{4}$$

In both of the two phases, the value of $a$ decreases from the initial value which is 2 to 0 until it reaches 0 at the end of the iterations. The range of the variable $r$ is between 0 and 1 which is a random number. The area of the whale where near the prey can be controlled by values of $A$ and $C$ vectors. By assigning values for $\vec{A}$ in the range [-1 and 1], the new location of the search agent can be identified between the current position of the whale and the best position.

Equation (5) is used to calculate the distance between the best position $\vec{X}^*(i)$ and the current position $X$, and it is also used to create a spiral-shaped approach.

$$\vec{X}(i+1) = e^{bk} \cdot \cos(2\pi k) \cdot \vec{D^*} + \vec{X^*}(i) \tag{5}$$

Where $D^*$ represents the distance between the whale and prey which is the best solution obtained so far.

$$\vec{D^*} = \left| \vec{X^*}(i) - \vec{X}(i) \right| \tag{6}$$

Where $b$ represents a constant value that identifies the logarithmic spiral shape and $k$ denotes a random number in the range [-1 and 1]. Forming the encircling shrinking mechanism and spiral-shaped mechanism, each mechanism has a 50% chance of being chosen through the iterations as shown in Equation (7).

$$\vec{X}(i+1) = \begin{cases} \overrightarrow{X^*} - \vec{A} \cdot \vec{D} & if\ p < 0{,}5 \\ e^{bk} \cdot \cos(2\pi k) \cdot \overrightarrow{D^*} + \overrightarrow{X^*}(i) & if\ p \geq 0{,}5 \end{cases} \tag{7}$$

Where $p$ is an arbitrary number between [0 and 1].

## 2.2 Searching for Prey

The exploration phase consists of random search techniques instead of updating the position according to the best position found. This strategy enhances the exploration phase. Finding prey depends on the techniques of changing the position of each whale. Therefore, $\vec{A}$ vector is used to control the whale to move far from the local whale. Throughout this phase, the position of whales is changing and it depends on the random search rather than the best position. This technique is resulted in performing global optima and overcoming local optima:

$$\vec{X}(i+1) = \overrightarrow{X_{rand}} - \vec{A} \cdot \vec{D} \tag{8}$$

$$\vec{D} = |\vec{C} \cdot \overrightarrow{X_{rand}} - \vec{X}\ | \tag{9}$$

Where $\overrightarrow{X_{rand}}$, is the position of one whale which is randomly chosen from the whales.

Algorithm 1 represents the WOA pseudo-code, and it can be noted that the population is initialized randomly. Then, the fitness of each search agent is evaluated. This process progresses until it reaches the best solution. After that, the coefficients variables are updated and a random number is used to update the position of agents using Equations (2) and (8) or Equation (5).

WOA can guarantee the convergence as it updates the position according to the best solution obtained. As a result, WOA may stick in the local optima and because of decreasing linearly from 2 to 0, $a$ is the main influence to balance on both phases.

## Algorithm 1: Whale Optimization Algorithm

Explanation detail of WOA:
Initialize population of whales, $X_n$ where ($n= 1,2,3,.....,m$)
Evaluate the fitness function of each agent
$X^*$= find the best search agent (whale)
  **While** (*iter < Maximum iterations*)
    **For each solution**
      Update *a, A, C, L,* and *p*
      **If1** *(p<0,5)*
        **If2** *(/A/ < 1 )*
          Update the location of the present search agent by Eq. (1)
        **Else if2***(/A/ ≥ 1)*
          Randomly search agent selected (*Xrand*)
          Update the location of the current search agent by the Eq. (7)
        **End if2**
      **Else if1** *(p ≥ 0,5 )*
        Update the location of the current search agent by the Eq. (5)
      **End if1**
    **End for**

Return search agents if it goes beyond the search space
Find fitness value for search agents
Update $X*$ if there is a better solution *iter=iter +1*
*End while*
Return $X*$

## 2.3 WOA Modifications and Hybridizations

Different types of modifications have been proposed since 2016. Table 1 illustrates the essential modifications of WOA. WOA has been hybridized with different metaheuristic algorithms. Therefore, Table 2 presents several WOA hybridizations.

**Table 1 WOA Modifications**

| Modification name | Reference | Purpose | Conclusion |
|---|---|---|---|
| **WOA in neural networks** | 2018, [34] | WOA is used as an optimizer to control weight and biases in neural networks. | Results presented that neural network by using WOA performs better compared to the Backpropagation algorithm. |
| **Chaotic WOA** | 2018, [35] | Chaos was used to control the status of WOA and to improve the performance of convergence speed, and achieve a better result. | Ten maps were tested in order to develop a chaotic set. CWOA improved the efficiency of WOA and balances between exploration and exploitation by using 0.7 as an initial point. |
| **Memetic WOA** | 2018, [22] | Avoiding local optima is a drawback of WOA. Therefore, MWOA was proposed in order to prevent WOA from this problem. | MWOA added a chaotic search embedded inside the exploration phase and creates stability between exploration and exploitation. |
| **ILWOA** | 2018, [36] | The decreasing cloud physical machine number was the aim of improving ILWOA due to the available bandwidth. | ILWOA was tested on 25 mathematical functions and then the result compared to WOA. The result showed that ILWOA improved WOA performance. |
| **IWOA** | 2017, [37] | The control parameter $a$ is linear, so, it cannot work well with nonlinear problems inside the search process. Therefore, IWOA used some nonlinear strategies to overcome this problem. | The result showed that IWOA performed well compared to standard WOA in convergence speed. |

**Table 2 WOA Hybridizations**

| WOA Hybridization with | Reference | Purpose | Conclusion |
|---|---|---|---|
| **BAT** | 2019, [28] | Improving the exploration of WOA and obtaining a better solution in the exploitation phase was the aim of WOA-BAT. | The WOA-BAT improved the quality of results against standard WOA and other algorithms. So, WOA-BAT outperformed WOA and |

| | | | other competitive metaheuristic algorithms. |
|---|---|---|---|
| **Artificial neural network based on WOA** | 2018, [38] | Using WOA to overcome the balancing difficulties related to parameter settings. | Results of the neural network based on WOA showed better performance, which is 9.9% accuracy. |
| **PSO** | 2018, [27] | The aim of PSO-WOA was to obtain better results for solving numerical functions that are global. | PSO embedded inside the hunting phase and the result was more efficient compared to the standard WOA. |
| **BS (Brain Storm)** | 2018, [39] | Privacy is a big challenge in cloud computing, so the secret key of data was identified by BS-WOA. | Results showed that BS-WOA obtained better security by protecting the confidentiality and effectiveness of data in the cloud. |
| **CBO (colliding bodies optimization)** | 2017, [25] | The aim of WOA-CBO was to improve the accuracy result, reliability and convergence speed. | WOA-CBO compared with the standard WOA and results showed that WOA-CBO performed better than WOA. |
| **MFO** | 2018, [40] | Avoiding time-consuming for determining the best optimal thresholding in multi-threshold was the aim of WOA-MFO. | WOA-MFO was compared with five algorithms. As a result, WOA-MFO showed a better result in terms of speed, the best fitness value, and the ANOVA test. |
| **LS (Local Search)** | 2018, [24] | Reducing computational cost and avoiding local optima. | The best result could be achieved quickly by using various techniques, for example, swamp mutation, local search strategy, and insert-reversed block. |

## 3. Grey Wolf Optimization

GWO was proposed in [17], which is motivated by the idea of hunting mechanism and hierarchy level among grey wolves in wildlife. The grey wolves are classified into four categories in GWO, namely; alpha ($\alpha$) wolf leader, beta ($\beta$) helping the leader, delta ($\delta$) follows both previous wolves and omega ($\omega$) [17]. Figure 1 shows the grey wolves' hierarchy.

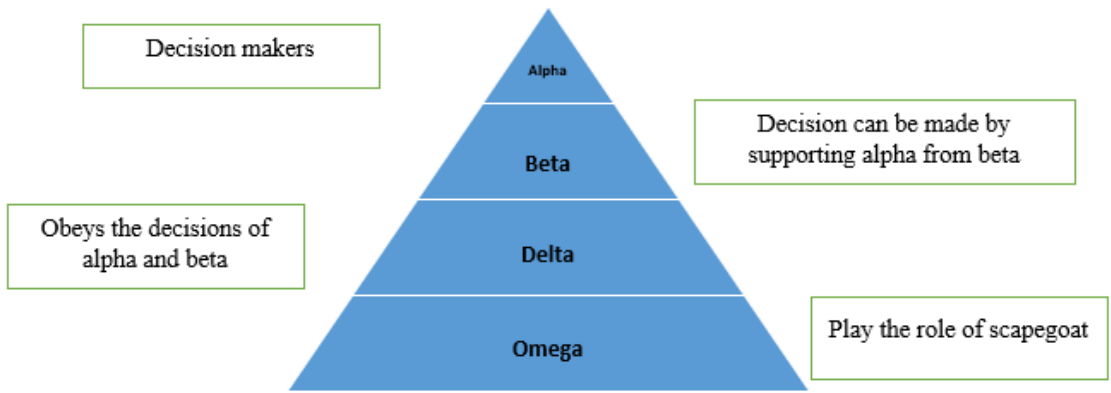

**Figure 1 Grey Wolves Hierarchy**

GWO has a social hierarchy, the first best solution is alpha ($\alpha$), then the beta ($\beta$), and the third-best solution is the delta ($\delta$). The remaining candidates' solution is called omega ($\omega$). These wolves ($\omega$) follows the other three wolves, which are above omega ($\omega$) in the hierarchy.

## 3.1 Encircling prey

Grey wolves try to encircle the prey in order to hunt by using Equations (10) and (11).

$$\vec{X}(i+1) = \overrightarrow{Xl}(i) - \vec{A} \cdot \vec{D} \tag{10}$$

$$\vec{D} = \left| \ \vec{C} \cdot \ \overrightarrow{Xl}(i) - \vec{X}(i) \ \right| \tag{11}$$

Where $\overrightarrow{Xl}(i)$ denotes the location of the prey at iteration $i$. $\vec{X}(i+1)$ is the location of a grey wolf. $A$ and $C$ are coefficient vectors, which can be calculated as follows:

$$\vec{A} = 2 \cdot \vec{a} \cdot \vec{r_1} + \vec{a} \tag{12}$$

$$\vec{C} = 2 \cdot \vec{r_2} \tag{13}$$

Decreasing $a$ value from 2 to 0 is happening in both phases until GWO reaches the maximum iteration. $\vec{r_1}$ and $\vec{r_2}$ are random numbers in the range of [0,1]. The area of wolves where near the prey can be controlled by the values of $A$ and $C$ vectors.

## 3.2 Hunting

After the encircling mechanism, a grey wolf starts to hunt the best solution. Despite the fact that the best solution required to be optimized, so, alpha wolf stores the best solution in each iteration and it changes if the solution is improved. The location of the prey can be identified by Beta and delta. Thus, the best solutions are saved by each type of grey wolves and used to update the position of grey wolves by using the following equations.

$$\overrightarrow{D\alpha} = \left| \ \vec{C_1} \cdot \ \overrightarrow{X\alpha} - \vec{X} \ \right|,$$

$$\overrightarrow{D_\beta} = \left| \ \vec{C_2} \cdot \ \overrightarrow{X_\beta} - \vec{X} \ \right|, \tag{14}$$

$$\overrightarrow{D_\delta} = \left| \ \vec{C_3} \cdot \ \overrightarrow{X_\delta} - \vec{X} \ \right|$$

$$\overrightarrow{X_1} = \overrightarrow{X_\alpha} - \overrightarrow{A_1} \cdot \overrightarrow{D_\alpha} \ , \overrightarrow{X_2} = \overrightarrow{X_\beta} - \overrightarrow{A_2} \cdot \overrightarrow{D_\beta} \ ,$$

$$\overrightarrow{X_3} = \overrightarrow{X_\delta} - \overrightarrow{A_3} \cdot \overrightarrow{D_\delta} \tag{15}$$

$$\vec{X}(i+1) = \frac{\overrightarrow{X_1} + \overrightarrow{X_2} + \overrightarrow{X_3}}{3} \tag{16}$$

## 3.3 Attacking Prey (Exploitation)

Hunting mechanism can be done by a grey wolf, which tries to stop the movement of the prey in order to attack them in this step. This mechanism is done by declining the value of $a$. The value of $\vec{A}$ is also reduced by the value $a$ and it is in the range of [-1,1]. Attacking the prey can be done by a grey wolf, if $\vec{A}$ is greater than -1 and less than 1. However, GWO suffers from stagnation in the local optima and researchers are trying to discover different mechanisms to solve this problem [17].

## 3.4 Search for Prey (Exploration)

Alpha, beta and delta influence the searching mechanism. These three categories are different from each other. Thus, they require a mathematical equation to converge and attack the prey. So, the value of $\vec{A}$ is between -1 and 1, if the value is greater than 1 or less than -1, the search agents are forced to diverge from the prey. In addition, if $\vec{A}$ greater than 1, then the search agent tries to find better prey. $\vec{C}$ is another component factor, which influences the exploration phase in GWO.

Overall, the random population is created in the GWO algorithm. Alpha, beta, and delta assume the location of the prey. Then, the candidate solution distance is updated. After that, $a$ $is$ reduced from 2 to 0 to balance between both of the phases. Next, the search agents go away from attacking the prey, if $\vec{A}>1$. If $\vec{A}<1$, then, they go forward the prey. Finally, the GWO has reached a satisfactory result and is terminated. Algorithm 2 describes the detail of the GWO Algorithm.

### Algorithm 2 Grey Wolf Optimization Algorithm

---

Grey wolf population initialization $Xn$ ( $n=1, 2, ....., m)$
Initialize variables such as $a, A,$ and $C$
Evaluate the fitness of search agent
$X\alpha$ = the best search agent
$X\beta$ = the second-best search agent
$X\delta$= the third best search agent
  **While** ($iter<$ Maximum iterations)
    **For** each search agent
      Update the location of the present search agent by Equation (16)
    **End For**
  Update $a, A,$ and $C$
  Search agents evaluated by a fitness
  Update $X_\alpha, X\beta$ $and$ $X\delta$
  $iter=iter+1$
**End while**
return X$\alpha$

---

## 3.5 GWO Modifications and Hybridizations

Many types of research have been done by researchers to modify and hybridize GWO. Therefore, Paper [41] described both modifications and the hybridization of GWO in detail. However, Table 3 and Table 4 briefly mentioned several crucial modifications and hybridizations of GWO in order to know that GWO has not been hybridized with WOA.

**Table 3 GWO Modifications**

| GWO modifications | Reference | Purpose | Conclusion |
|---|---|---|---|
| **Modified GWO** | 2019, [42] | MGWO was proposed to tune recurrent neural network parameters, which then used for classifying students' performance. | As a result, MGWO could find the best solution than other competitive models. MGWO has a greater impact to improve the result of a recurrent neural network. |
| **Chaotic GWO** | 2018, [43] | Increasing the convergence speed was the purpose of this modification by adding different chaotic methods. | Ten chaotic maps were used and the best one was chosen in order to use it with GWO. Therefore, the results showed that CGWO improved standard GWO and it was better than other algorithms. |
| **Binary GWO** | 2018, [44] | There are different large-scale problems. Unite commitment problem was one of those problems that could be solved by using BGWO. | BGWO was used to solve the unite commitment problem. Then, it was compared to the standard GWO and a variety of binary algorithms. The result showed that BGWO outperformed well compare to them. |
| **Intelligent GWO** | 2018, [45] | The aim of IGWO was at solving different problems in companies, which sale power in the energy market. Companies using different strategies to increase their profit but they have difficulties in predicting the information about the future energy price. | IGWO was tested on 22 benchmark functions. IGWO was compared with GWO, Oppositional GWO, and PSO. IGWO results showed superior compared to the other algorithms. |
| **GWO** | 2017, [46] | GWO was proposed to design the modular neural network architecture. The aim of this work was to improve the performance of the human recognition system. | Results are compared to GA and Firefly Algorithm (FA). The GWO outperforms well compared to GA and FA. |
| **Power GWO** | 2015, [47] | GWO was used to solve complex optimization problems based on Power local optimization approach, which was essential for clustering. | PGWO was tested on seven benchmark functions and tested on nine data sets for clustering. Results showed that PGWO performed well against the most recent algorithms. |

**Table 4 GWO Hybridizations**

| GWO hybridizations with | Reference | Purpose | Conclusion |
|---|---|---|---|
| Dragonfly (DA) | 2019, [48] | The renewable energy system has some problems, such as voltage deviation, power loss, and decreasing fuel cost. Therefore, the aim of this approach was to solve these problems. | The result of the hybridized approach showed that it was faster and improved its performance when the IEEE 30 bus system was used to test. |
| Recurrent neural network | 2019, [42] | The learning experience and forecasting outcome of the student's results was the aim of this hybridization. | Results proved that the hybridized system improved the forecasting task in terms of accuracy compared to other models. |
| Long Short Term Memory (LSTM) | 2018, [49] | The recurrent neural network has some drawbacks related to accuracy, convergence speed. In this work, the GWO was used to train the LSTM recurrent neural network. | Simulation results presented that GWO can improve the performance of the recurrent neural networks by training the LSTM recurrent neural networks. |
| Fireworks Algorithm (FWA) | 2018, [50] | The aim of this hybridization was to combine the two most efficient algorithms, which have been inspired by physics and nature. | The FWA-GWO was tested on 22 benchmarks functions and then compared to FWA and GWO. Results showed that FWA-GWO outperformed the other two standard algorithms. |
| Flower Pollination Algorithm (FPA) | 2017, [51] | Hybridizing both algorithms to have a better solution in solving real-world applications was the aim of this hybridization. | The hybridized approach was verified on 6 benchmark functions and then compared against PSO, FPA, and GWO. So, GWO-FPA showed superiority in its performance. |
| Sine Cosine Algorithm (SCA) | 2017, [52] | To improve the quality solution of GWO, GWO was hybridized with SCA. | The results were compared with standard GWO, SCA, WOA, ALO, and PSO. It can be said that GWO-SCA performed well in solving test functions and solving real-world problems. |
| GA | 2016, [53] | Solving the economic dispatch problems was the aim of this approach. | GWO was hybridized with a crossover and mutation mechanism for improving the performance. The results showed equality in some cases and better results in others. |

# 4. Our Approach: WOAGWO

Based on the previous sections about WOA and GWO, the proposed approach is explained in this section by combining WOA and GWO to enhance the performance of WOA in terms of efficiency in exploitation phase to obtain better solutions

In general, the standard WOA can perform well in finding the best solution. However, refining the optimum solution in each iteration is not sufficient. Therefore, WOA is hybridized with GWO in order to improve the performance of WOA. The hybridized algorithm is called WOAGWO. As a result, the standard WOA is hybridized by adding two sections. Firstly, we added a condition inside the exploitation phase in WOA for improving the hunting mechanism. According to Eq. (16), *A1, A2, and A3* have a greater impact on exploitation performance. Therefore, a new condition is added to the standard exploitation phase of WOA for avoiding local optima where each *A* is less than 1 or greater than -1. Secondly, we adapted Equation (14), (15) and (16). And we used them inside the condition that was added to the exploitation phase which includes *A1, A2, and A3*. Finally, another new condition is added to the exploration phase to make the current solution move towards the best solution. It also avoids the whale to change to a position that is not better than the previous position.

The differences between WOAGWO and WOA are Equation (14), (15) and (16) which are added to the exploitation of WOA. A new mechanism is added inside the exploration phase to improve the solution. Therefore, this condition with equations of GWO improves the hunting mechanism of WOA. It also improves the best solution after each iteration and generates better performance regarding local optima. Furthermore, using the condition inside the exploration phase improves the searching capability as it improves the quality of the solution if it exists.

WOAGWO is started by initializing the population size of the search agents (which includes both whales and wolves). Then, the population goes through a process to amend the agents if they go beyond the search space. Therefore, the fitness function is calculated. If fitness is less than the Alpha_score (Best_Score, then Alpha_score is equal to fitness. After that, these variables are updated: *a, A, C, L,* and *p*. Then a random number is generated.

If the random number is less than 0,5, then it goes to another condition which is if ($/A/ < 1$). If this condition is true, then the new position is calculated using Eq.1. As a result, if the new position is better than the old position, then the old position is updated. However, if ($/A/ >=1$), then the new position is found using Eq. 2. Like the previous condition, the new position fitness is compared to the old fitness. If it is better than the old one, then the position is updated.

On the other hand, if the random value is greater than or equal to 1, then the new condition is counted which is if(($A1>-1 \parallel A1<1$). If these conditions are true, then the Alpha_position, Beta_position, and Delta_position are calculated using Eq. 15. Consequently, the new position is calculated by Eq. 16.

After the above steps, the new position requires checking either it is beyond the search space or not. If they are out of the feasible space, then the position is amended depending on the limitation. As a result, a new fitness value is calculated, and finally, the best fitness value is returned.

WOAGWO pseudocode and flowchart are presented in Algorithm 3 and Figure 2.

Algorithm 3 WOAGWO

---

Explanation of WOAGWO:
Initialize WOAGWO population $X_i$ where ($i= 1,2, 3,.....,n$)
Evaluate the fitness function for each search agent
$X*$= the best search agent
  **While** (*iter < Max_iterations*)
    **For each solution**
     Update *a, A, C, L,* and *p*
       **If1** (*p<0,5*)
         **If2** (*/A/ < 1* )
          Calculate new location of the present search agent by Eq. (1)
             **If3** (*fcurrent< fprevious*)
               *Position= new_Position.*
             **End if3**
         **Else if2**(*/A/ ≥ 1*)
          Randomly choose search agent ($X_{rand}$)
          Change the location of the present search agent by the Eq. (2)
             **If4** (*fcurrent< fprevious*)
               *Position= new_Position.*
             **End if4**
        **End if2**
       **Else if1** (*p ≥ 0,5* )
         **If5**((*A1>-1 || A1<1*)&&(*A2>-1 || A2<1*)&&(*A3>-1 || A3<1*))
            Location of the current search agent updated by Eq. (16)
         **End if5**
      **End if1**
     **Enf for**
      Return search agents to inside the search space if it goes beyond the search space
      Fitness value for each search agent is calculated
      Update $X^*$ if there is a better solution *iter=iter +1*
**End while**
Return $X^*$

---

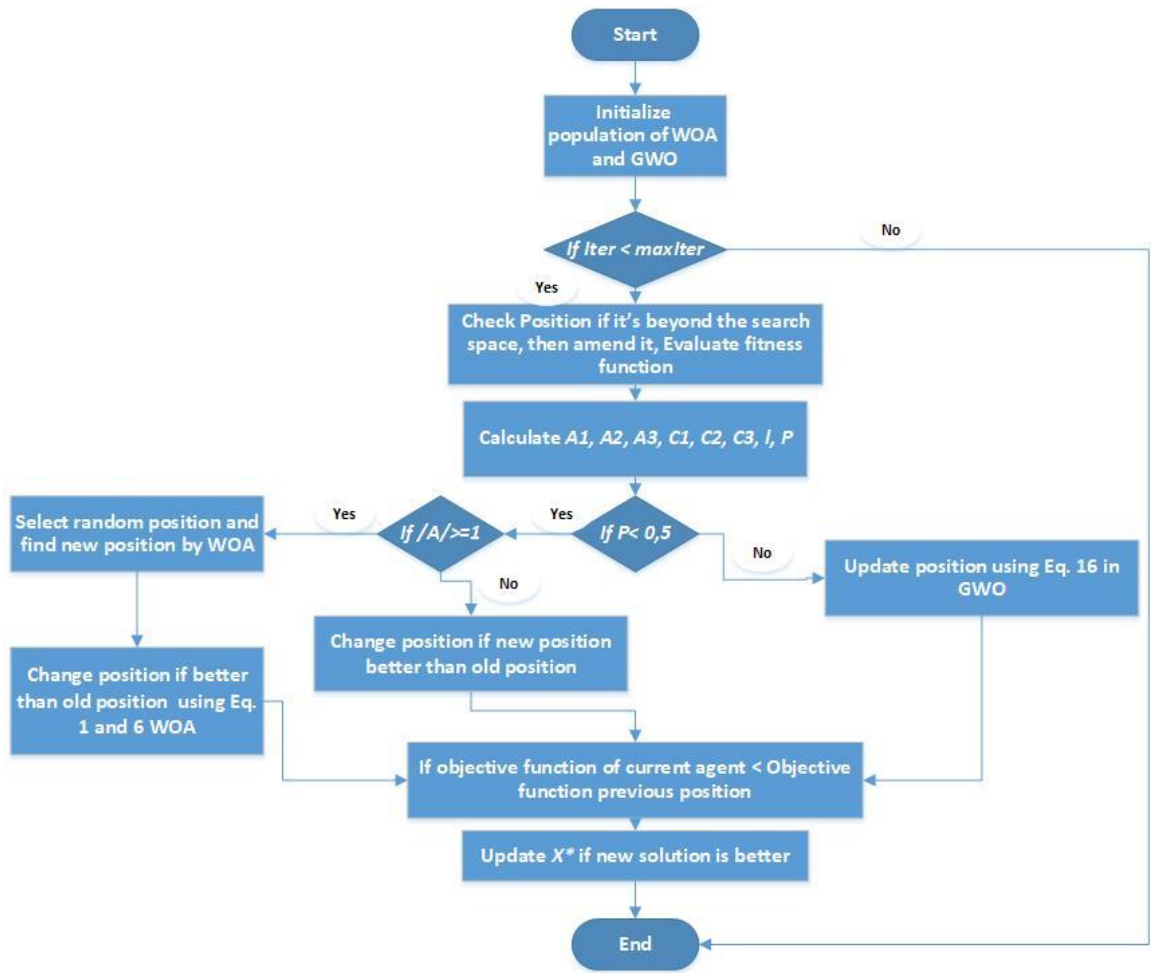

**Figure 2 WOAGWO Flowchart**

# 5. Experimental Result and Discussion

WOAGWO algorithm is implemented and evaluated against 23 benchmark functions [28], 25 benchmark functions from CEC2005, and 10 benchmark functions from CEC2019. The following subsections describe benchmark functions, experimental setup, evaluation criteria, statistical results, and evaluations of WOAGWO against other metaheuristic algorithms.

## 5.1 Benchmark Functions

Three various benchmark functions are conducted in order to verify our proposed WOAGWO. The first benchmark function is 23 functions. Then, the CEC2005 benchmark function is used. These are 25 functions of CEC2005. The third part of the benchmark functions is CEC2019. These functions include multimodal functions, unimodal functions, expanded multimodal functions, and hybrid composition functions. These benchmark test functions can be seen in [28].

## 5.2 Experimental Setup

The code is implemented by using Matlab R2017b on Windows 10. The first population is initialized randomly in order to have a better and accurate result. Table 5, shows parameter initialization for implementation.

**Table 5 Parameter Initialization**

| Number | Parameters | Times | Algorithms |
|---|---|---|---|
| **1** | Population size | 30 | WOAGWO, |
| **2** | Maximum iteration | 500 | WOA, and |
| **3** | Run time for each function | 30 | GWO |

## 5.3 Evaluation Criteria

Different ways are used for evaluating WOAGWO. The next is the evaluation points:

1) Presenting average and standard deviation.
2) Comparing WOAGWO with WOA.
3) Comparing WOAGWO with GWO.
4) Comparing WOAGWO with other metaheuristic algorithms (DE, ABC, BSO, and WOA).
5) Creating a Box and Whisker plot for comparison of WOA, GWO, and WOAGWO.

## 5.4 WOAGWO Vs. WOA

The performance of WOAGWO can be evaluated using these functions. Functions *f1-f7* are called unimodal functions, which have a single solution. As a result, the WOAGWO exploitation capability can be evaluated by using these unimodal functions. Table 6, shows that WOAGWO has better exploitation capability compared to the standard WOA in all seven functions.

In other functions, such as *f8-f23,* which are multimodal functions and they are useful to assess our proposed algorithm in terms of exploration. Table 6, shows that WOAGWO outperforms in 13 out of 16 multimodal functions. As a result, it can be said that WOAGWO improves the performance of WOA in exploration. Nonetheless, the WOAGWO algorithm has the same result as WOA for function 16. Conversely, WOA performs well in both functions *f16 and f17.*

**Table 6 Comparison of WOAGWO with WOA**

| F | WOA | | WOAGWO | | GWO | |
|---|---|---|---|---|---|---|
| | avg | std | avg | std | avg | std |
| 1 | 1.2E-74 | 5.9431E-74 | **0** | 0 | 1.94E-27 | 3.2433E-27 |
| 2 | 2.37E-51 | 8.8634E-51 | **1.29E-210** | 0 | 7.4743E-17 | 4.4536E-17 |
| 3 | 50945 | 14806.2448 | **0** | 0 | 9.2636E-06 | 1.6703E-05 |
| 4 | 52.426 | 24.6188925 | **1.29E-213** | 0 | 8.4924E-07 | 9.4388E-07 |
| 5 | 28.02927 | 5.04953113 | **1.20E-08** | 2.9005E-08 | 27.037 | 0.8268246 |
| 6 | 0.4356 | 0.20393211 | **1.80E-11** | 2.5159E-11 | 0.772 | 0.35489112 |
| 7 | 0.0026 | 0.00202219 | **2.08E-04** | 0.00015661 | 0.0021 | 0.0010843 |
| 8 | -10424 | 2699.73123 | **-1.26E+04** | 5.4772E-05 | -5834.2 | 1166.4211 |
| 9 | 1.89E-15 | 1.0477E-14 | **0** | 0 | 3.2605 | 3.79275955 |
| 10 | 4.8E-15 | 2.3118E-15 | **8.88E-16** | 4.0117E-31 | 1.0629E-13 | 2.1811E-14 |
| 11 | 0.011 | 0.04289421 | **0** | 0 | 0.0043 | 0.00768593 |
| 12 | 0.02 | 0.00971086 | **3.93E-12** | 9.5317E-12 | 0.0431 | 0.01889672 |
| 13 | 0.5672 | 0.29065488 | **5.07E-11** | 9.101E-11 | 0.6543 | 0.24743806 |
| 14 | 3.258 | 3.18587253 | **0.998** | 4.5168E-16 | 4.5917 | 3.94839187 |
| 15 | 0.000566 | 0.00036856 | **3.14E-04** | 2.7985E-05 | 0.0038 | 0.00740239 |
| 16 | **-1.0316** | 0.18834353 | **-1.0316** | 6.7752E-16 | **-1.0316** | 6.6613E-16 |

| 17 | **0.3979** | 0.07264355 | 0.398 | 0.00050379 | **0.3979** | 4.2687E-06 |
| 18 | **3** | 0.54771826 | 3.0001 | 0.00012058 | **3** | 5.9628E-05 |
| 19 | -3.856 | 0.70595763 | **-3.8612** | 0.00248348 | -3.8613 | 0.00270342 |
| 20 | -3.225 | 0.61610521 | **-3.2677** | 0.06883349 | -3.2435 | 0.08792772 |
| 21 | -8.746 | 2.94478365 | **-10.1532** | 3.6134E-15 | -10.1516 | 0.0009222 |
| 22 | -7.6138 | 3.30341493 | **-10.4028** | 9.0336E-15 | -10.401 | 0.00152396 |
| 23 | -6.7571 | 3.90115355 | **-10.5363** | 0 | -10.3545 | 0.97046316 |

CEC2005 benchmark function also used to evaluate the WOAGWO algorithm. Table 7, illustrates that WOAGWO exploitation performance is better than WOA in *f2, f3, f3, f4, and f5*. However, WOA performs well only in *f1*. To evaluate exploration capability, *f6-f12* is used. As a result, WOAGWO performs well in all functions except *f7,* which has the same result as WOA. Despite having worse results in 4 functions compared to WOA, WOAGWO performs well in 10 out of 14 functions. Overall, we can say that WOAGWO improves WOA in exploration and exploitation in 19 functions, WOA is better in 4 functions and they are the same in 1 function.

WOAGWO also compared with GWO in Table 7 that shows WOAGWO performs better than GWO in 4 out of 5 unimodal functions. However, WOAGWOA exploration performance improves only in 4 multimodal functions. WOAGWO is also better than WOA in 9 functions. In general, WOAGWO is efficient in 16 functions while GWO is better than WOAGWO in 8 functions and they are the same in *f7*.

Overall, WOAGWO has better functionality in 14 benchmark functions compared to WOA and GWO. It has the same result as WOA and GWO in 1 function. However, GWO has a better result in 7 functions while WOA performs well only in 3 test functions.

**Table 7 WOA, WOAGWO, and GWO Comparison Results on CEC2005**

| F | WOA | | WOAGWO | | GWO | |
|---|---|---|---|---|---|---|
| | avg | std | avg | std | avg | std |
| **1** | **1.64E-07** | 3.26E-07 | 90.8169 | 125.6311 | 7.54E+01 | 125.2783 |
| **2** | 1.24E+04 | 3.93E+03 | **528.9248** | 677.455 | 572.4044 | 838.9729 |
| **3** | 5.32E+06 | 5.61E+06 | **2.09E+06** | 2308557 | 2.16E+06 | 5018469 |
| **4** | 2.10E+04 | 8.64E+03 | **1.29E+03** | 1406.165 | 1.37E+03 | 1394.573 |
| **5** | 3.56E+03 | 3.15E+03 | **619.1757** | 1145.875 | 805.3676 | 1911.879 |
| **6** | 6.75E+05 | 2.68E+05 | 6.57E+05 | 1963062 | **5.66E+05** | 1975092 |
| **7** | **1.27E+03** | 6.3685 | **1.27E+03** | 0.136957 | **1.27E+03** | 0.088189 |
| **8** | 20.414 | 0.0984 | **20.2854** | 0.148188 | 20.4805 | 0.096684 |
| **9** | 44.5728 | 17.454 | **16.7816** | 9.367935 | 17.0284 | 9.166468 |
| **10** | 71.1216 | 20.5328 | 28.5578 | 14.38448 | **25.2281** | 13.05085 |
| **11** | 9.12 | 1.4296 | 4.682 | 1.856523 | **4.2209** | 1.113957 |
| **12** | 1.62E+04 | 1.85E+04 | **3.88E+03** | 4017.933 | 4.32E+03 | 6233.134 |
| **13** | 4.1037 | 1.8862 | **1.6063** | 0.891403 | 1.6297 | 0.717426 |
| **14** | 3.9046 | 0.2729 | 3.3739 | 0.336788 | **3.1787** | 0.530905 |
| **15** | 20.6418 | 28.8754 | **19.0478** | 25.45496 | 20.648 | 28.87942 |
| **16** | 48.4241 | 96.6469 | 55.5494 | 51.0079 | **35.949** | 46.76869 |

| | | | | | |
|---|---|---|---|---|---|
| **17** | 44.4717 | 76.5331 | 62.6943 | 44.52695 | **44.3661** | 51.65567 |
| **18** | 293.6006 | 140.8512 | **47.2492** | 85.71267 | 274.6266 | 123.9029 |
| **19** | 300.7357 | 116.2018 | **200.0032** | 0.002584 | 300.0013 | 58.72239 |
| **20** | 193.3517 | 172.063 | **160.0176** | 81.33856 | 319.5926 | 57.3539 |
| **21** | **266.855** | 182.8915 | 332.0461 | 170.8737 | 345.2527 | 183.325 |
| **22** | 290.3025 | 120.8642 | **254.049** | 55.26927 | 260.003 | 81.36783 |
| **23** | **225.9111** | 223.3001 | 481.6775 | 140.5439 | 484.306 | 187.9177 |
| **24** | 200 | 5.65E-07 | **167.9354** | 43.36581 | 200.0002 | 0.000205 |
| **25** | 197.9812 | 5.65E-07 | 111.08 | 12.35356 | **106.2192** | 10.39683 |

CEC2019 is also used to test WOAGWO and compared it with WOA and GWO. Table 8 and Figure 3 show that WOAGWO is better than WOA in seven functions, such as *f1, f2, f4, f5, f7, f8,* and *f9* and it has the same result as WOA in *f3*. However, WOA is better in *f6 and f10*.

Comparing WOAGWO against GWO as shown in Table 8. WOAGWO performs well in five functions; they have the same result in both functions (*f2, f8*). However, GWO is better than WOAGWO in 3 functions.

Overall, WOAGWO is better than WOA and GWO in 5 multimodal benchmark functions and two functions have the same results. WOA is better than WOAGWO in 2 functions. Finally, GWO is better than WOA and WOAGWO in 1 function.

**Table 8 WOA, WOAGWO, and GWO Comparison Results on CEC2019.**

| F | WOA | | WOAGWO | | GWO | |
|---|---|---|---|---|---|---|
| | **avg** | **std** | **avg** | **std** | **avg** | **std** |
| **1** | 2.10E+10 | 3.57E+10 | **4.76E+04** | 5186.077 | 2.13E+08 | 3.07E+08 |
| **2** | 1.84E+01 | 1.61E-02 | 18.3441 | 0.000472 | **1.83E+01** | 0.000304 |
| **3** | **1.37E+01** | 7.23E-15 | 13.7024 | 1.83E-05 | **1.37E+01** | 1.922208 |
| **4** | 3.48E+02 | 1.72E+02 | 253.6765 | 538.7369 | 3.01E+02 | 686.8153 |
| **5** | 3.03E+00 | 4.86E-01 | **2.4257** | 0.262064 | 2.43E+00 | 0.251607 |
| **6** | **1.03E+01** | 1.39E+00 | 11.3655 | 1.641948 | 1.19E+01 | 0.730745 |
| **7** | 6.14E+02 | 2.98E+02 | 587.6149 | 348.9018 | **5.35E+02** | 292.0204 |
| **8** | 6.03E+00 | 5.66E-01 | **5.587** | 1.022585 | **5.40E+00** | 0.993956 |
| **9** | 5.93E+00 | 6.85E-01 | **5.6705** | 0.880983 | 1.47E+01 | 49.95142 |
| **10** | **2.13E+01** | 1.35E-01 | 21.5576 | 0.092245 | 2.15E+01 | 0.068513 |

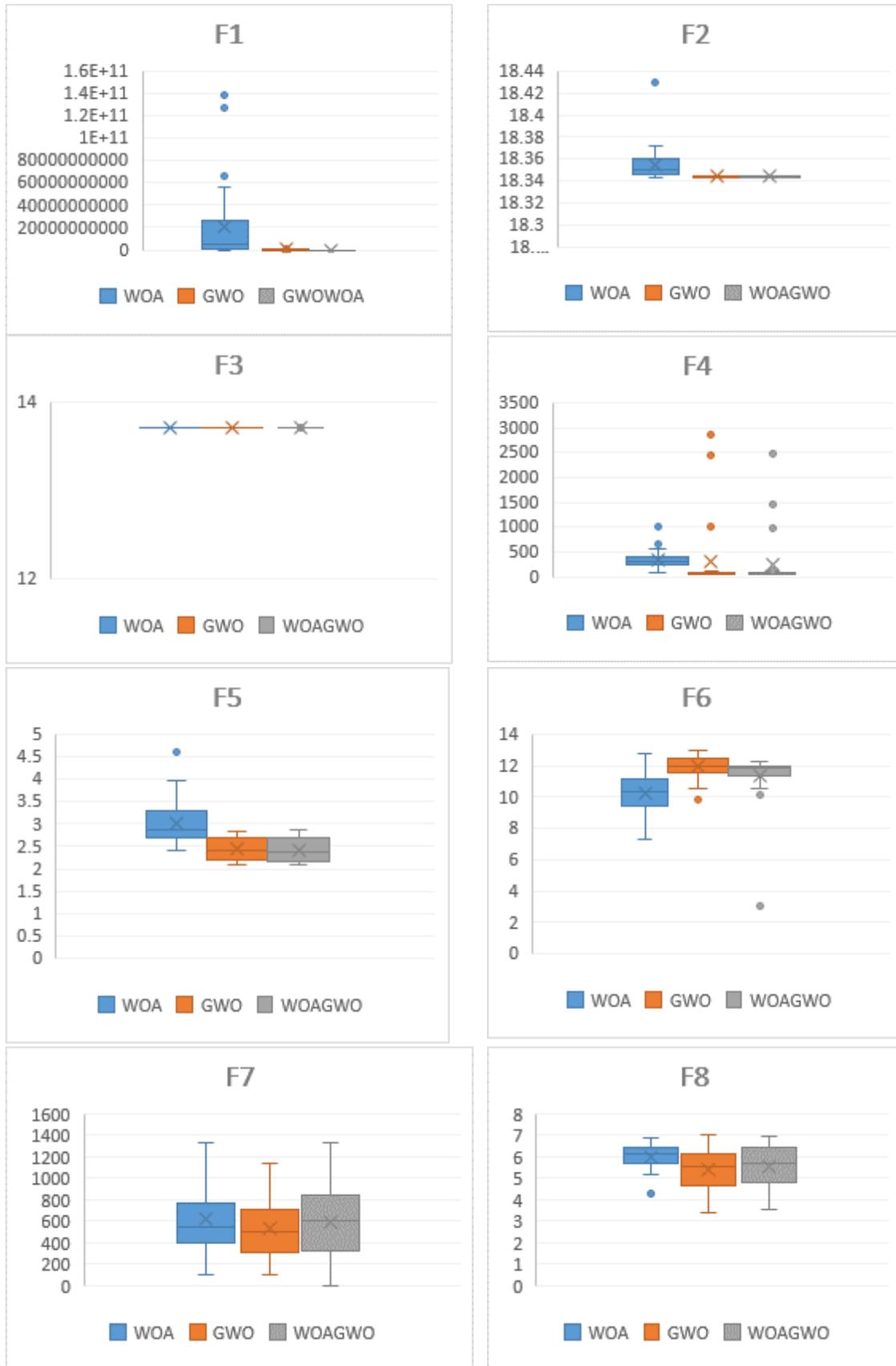

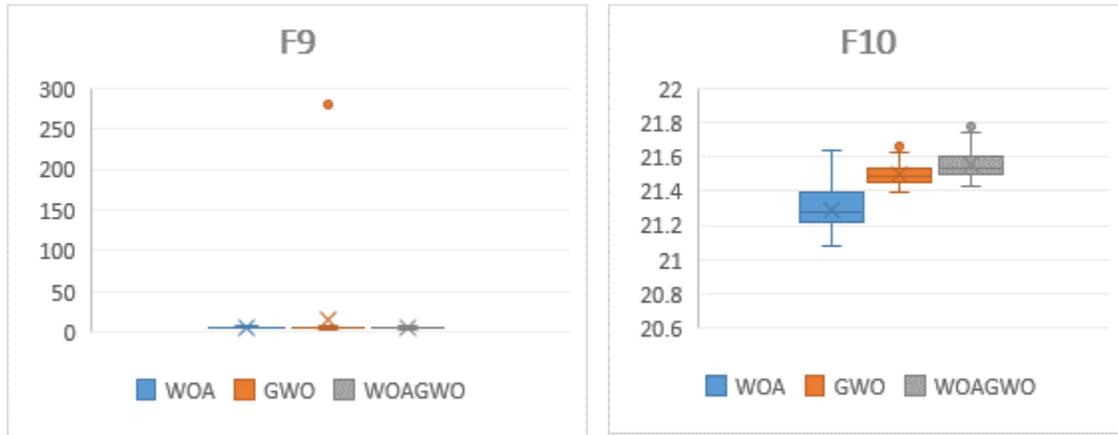

**Figure 3 Box and Whisker plot of WOA, GWO, and WOAGWO on CEC2019.**

## 5.5 Statistical Test

In order to show that the results are either significant or not in Table 6, Table 7, and Table 8, the Wilcoxon rank-sum test is used to find the $p$ values for all benchmark test functions. The results of the Wilcoxon rank-sum test are shown in Table 9. The $p$-value found for all the benchmark functions and for each of the above-mentioned tables. The $p$-value obtained between WOA vs. WOAGWO.

Table 9, shows that WOAGWO obtained significant results against WOA in all unimodal and multimodal functions except functions 9 and 11 for the first column, which is 23 function results. However, WOAGWO does not have significant value in 6 functions while it is compared to WOA.

By obtaining the $p$-value from comparing WOAGWO against WOA, Table 9, shows that WOAGWO has significant results in 13 functions out of 25 functions from CEC2005.

In addition, WOAGWO shows that the $p$-value of CEC2019 test functions, it can be seen from Table 9 that WOAGWO obtained a significant result in 6 out of 10 functions against WOA.

The reasons behind these results as shown in Table 9 are WOA has a crucial technique to update the whale position in the exploration phase. GWO has very effective performance when it is used inside WOAGWO for the exploitation. GWO has a great impact on improving the performance of WOAGWO over WOA and GWO since Beta and Delta types of wolves save the best solutions. These solutions are used to update the position of the whale inside the WOAGWO. The other reason is that decreasing the value of $a$ in the range of [-1,1] increases the capability of whales to attain the best solution in each iteration. These reasons have a significant impact on WOAGWO over the original WOA and GWO when tested on 23 classical benchmark functions, CEC2005, and CEC2019.

**Table 9 $P$-Value of WOAGWO against WOA for 23 Benchmark Functions, CEC2005, CEC2019.**

| F | 23 Functions | CEC2005 | CEC2019 |
|---|---|---|---|
| **1** | <u>0</u> | 1 | <u>0</u> |
| **2** | <u>0</u> | <u>0</u> | <u>0</u> |
| **3** | <u>0</u> | <u>1.92E-09</u> | 1 |
| **4** | <u>0</u> | <u>0</u> | <u>3.1E-14</u> |
| **5** | <u>0</u> | <u>2.57E-08</u> | <u>0</u> |
| **6** | 0 | 0.099275 | 1 |
| **7** | 0 | 0.999957 | 0.263969 |
| **8** | 0 | 2.55E-09 | 0.017695 |
| **9** | 1 | 0 | 0.032394 |
| **10** | 0.001612662 | 0 | 1 |
| **11** | 1 | 0 | |
| **12** | 0 | 5.86E-09 | |
| **13** | 0 | 0 | |
| **14** | 0.080371993 | 0 | |
| **15** | 0 | 1 | |
| **16** | 1 | 1 | |
| **17** | 1 | 1 | |
| **18** | 1 | 1.73E-08 | |
| **19** | 1.19485E-07 | 0.013997 | |
| **20** | 1.15567E-08 | 0.583317 | |
| **21** | 0 | 1 | |
| **22** | 0 | 0.964067 | |
| **23** | 0 | 1 | |
| **24** | | 1 | |
| **25** | | 0.163099 | |

## 5.6 Comparing WOAGWO with Hybrid and Metaheuristic Algorithms

WOAGWO as the hybrid algorithm is compared with the WOA-BAT algorithm by using the CEC2019 test functions. The results of WOA-BAT is obtained from [18]. Table 10, presents that WOAGWO performs well in 6 out of 10 functions. This means that using GWO hunting techniques in the exploitation phase of WOA is the reason behind the achieved result. Though, WOA-BAT had improved WOA. But, WOAGWO achieves the best results. It is believed that WOAGWO performance better than WOA-BAT.

**Table 10 Comparison Results of WOA_BAT and WOAGWO using CEC2019.**

| F | WOA-BAT | | WOAGWO | |
|---|---|---|---|---|
| | avg | std | avg | std |
| **1** | 7.60E+07 | 4.16E+08 | **4.76E+04** | 5186.077 |
| **2** | 1.75E+01 | 1.21E-01 | 18.3441 | 0.000472 |
| **3** | 1.27E+01 | 9.53E-04 | 13.7024 | 1.83E-05 |

| 4 | 2.12E+03 | 1.01E+03 | **253.6765** | 538.7369 |
|---|----------|----------|--------------|----------|
| 5 | 2.44E+00 | 6.67E-01 | **2.4257** | 0.262064 |
| 6 | 1.11E+01 | 1.55E+00 | 11.3655 | 1.641948 |
| 7 | 6.06E+02 | 3.90E+02 | **587.6149** | 348.9018 |
| 8 | 5.72E+00 | 7.18E-01 | **5.587** | 1.022585 |
| 9 | 2.28E+01 | 4.92E+01 | **5.6705** | 0.880983 |
| 10 | 2.12E+01 | 2.26E-01 | 21.5576 | 0.092245 |

In addition, different metaheuristic results are presented in this section, which is obtained from CEC2005. These results are taken from various optimization algorithms, such as DE, ABC, BSO, WOA, and WOAGWO. Table 11, illustrates that each algorithm is better than other algorithms in a different number of functions out of 25 functions. The following points represent the conduct of each algorithm on the number of functions:

- GA does not achieve the best results.
- DE obtained the best results in 3 functions out of 25.
- BSO attained well in 9 out of 25 functions.
- WOA takes the best results in 4 out of 25 functions.
- WOAGWO achieves the best results in 9 out of 25 functions.

As a result, each WOAGWO and BSO achieves best results in 9 out of 25 functions. Therefore, WOAGWO and BSO are better than the other three algorithms in 9 benchmark functions. WOAGOW is better than other algorithms in 2 unimodal functions and it is better in 7 hybrid benchmark functions. As a result, WOAGWO has sufficient capability of balancing between exploration and exploitation. In addition, WOA performs well in 4-hybrid benchmark functions. WOAGWO improves the performance of WOA from 4 to 9 functions in balancing exploration and exploitation.

However, BSO is better than the other algorithms in three unimodal functions, which means that BSO performs well in exploitation capability. BSO is also performed well in three multimodal functions. Therefore, the exploration performance of BSO is worse compared to WOAGWO.

DE has the third rank in comparison with the other algorithms in Table 11. It performs well in 2 unimodal functions and multimodal functions. However, ABC results have worse results compared to others. Finally, it can be said that WOAGWO is better than BSO, WOA, ABC, and DE in balancing between exploitation and exploration.

Overall, it can be said that WOAGWO is very competitive against DE, ABC, BSO, and WOA. WOAGWO performs well in 9 functions while BSO is better in 9 functions as well. Therefore, WOAGWO could improve the performance of WOA from 4 functions to 9 functions because of adding a conditioning technique inside the exploration phase to improve solution quality and adding the second condition inside the exploitation phase, which focuses on the $A$ value, improves the exploitation capability of WOAGWO. Furthermore, adapting Equations (14), (15), (16) and (17) improves the performance of WOA as it can be seen in Table 11, which shows that WOA is better than BSO only in 4 functions.

**Table 11 ABC, DE, BSO, WOA, and WOAGWO Comparison Results on CEC2005.**

| F | ABC | | DE | | BSO | | WOA | | WOAGWO | |
|---|-----|-----|-----|-----|-----|-----|-----|-----|--------|-----|
| | avg | std | avg | std | avg | std | avg | std | avg | std |
| 1 | 2.20E−02 | 4.08E−02 | 1.79E−04 | 1.31E−04 | **-4.50E+02** | 3.50E-14 | 8.83E+00 | 3.26E−07 | 9.08E+01 | 125.6311 |
| 2 | 2.73E+04 | 4.05E+03 | 2.12E+02 | 9.29E+01 | **-4.48E+02** | 9.36E-01 | 1.09E+04 | 3.93E+03 | 5.29E+02 | 677.455 |
| 3 | 1.22E+08 | 2.90E+07 | 6.28E+06 | 2.09E+06 | **2.04E+06** | 7.23E+05 | 3.02E+06 | 5.61E+06 | 2.09E+06 | 2308557 |
| 4 | 3.38E+04 | 4.49E+03 | **1.15E+03** | 7.23E+02 | 2.78E+04 | 8.05E+03 | 1.83E+04 | 8.64E+03 | 1.29E+03 | 1406.165 |
| 5 | 8.30E+03 | 8.00E+02 | **5.63E+02** | 2.84E+02 | 4.70E+03 | 1.22E+03 | 2.87E+03 | 3.15E+03 | 6.19E+02 | 1145.875 |
| 6 | 3.65E+05 | 2.58E+05 | **3.94E+01** | 2.98E+01 | 1.26E+03 | 9.48E+02 | 1.39E+05 | 2.68E+05 | 6.57E+05 | 1963062 |
| 7 | 4.89E+03 | 2.88E+01 | 4.70E+03 | 9.01E−11 | **6.25E+02** | 3.25E+02 | 1.27E+03 | 6.3685 | 1.27E+03 | 0.136957 |
| 8 | 2.10E+01 | 6.86E−02 | 2.10E+01 | 7.75E−02 | **-1.20E+02** | 9.90E-02 | 2.03E+01 | 0.0984 | 2.03E+01 | 0.148188 |
| 9 | 2.10E+02 | 1.35E+01 | 1.46E+02 | 2.87E+01 | **-2.86E+02** | 1.27E+01 | 4.22E+01 | 17.454 | 1.68E+01 | 9.367935 |
| 10 | 2.46E+02 | 9.04E+00 | 2.15E+02 | 1.13E+01 | **-2.93E+02** | 8.79E+00 | 6.23E+01 | 20.5328 | 2.86E+01 | 14.38448 |
| 11 | 4.05E+01 | 1.37E+00 | 4.04E+01 | 1.35E+00 | 1.10E+02 | 2.51E+00 | 8.87E+00 | 1.4296 | **4.68E+00** | 1.856523 |
| 12 | 4.02E+05 | 5.17E+04 | 1.82E+04 | 1.19E+04 | 2.84E+04 | 1.99E+04 | 1.60E+04 | 1.85E+04 | **3.88E+03** | 4017.933 |
| 13 | 2.31E+01 | 1.45E+00 | 1.79E+01 | 1.49E+00 | **-1.26E+02** | 1.05E+00 | 4.42E+00 | 1.8862 | 1.61E+00 | 0.891403 |
| 14 | 1.36E+01 | 1.34E−01 | 1.37E+01 | 1.32E−01 | **-2.87E+02** | 3.78E-01 | 3.92E+00 | 0.2729 | 3.37E+00 | 0.336788 |
| 15 | 3.06E+02 | 5.76E+00 | 2.70E+02 | 9.66E+01 | 5.43E+02 | 7.94E+01 | 2.19E+01 | 28.8754 | **1.90E+01** | 25.45496 |
| 16 | 2.63E+02 | 9.94E+00 | 2.54E+02 | 4.05E+01 | 2.87E+02 | 1.34E+02 | **3.35E+01** | 96.6469 | 5.55E+01 | 51.0079 |
| 17 | 2.86E+02 | 1.72E+01 | 2.81E+02 | 4.62E+01 | 3.10E+02 | 1.57E+02 | **2.29E+01** | 76.5331 | 6.27E+01 | 44.52695 |
| 18 | 9.60E+02 | 5.84E+00 | 9.06E+02 | 7.56E−01 | 9.17E+02 | 1.36E+00 | 2.94E+02 | 140.8512 | **4.72E+01** | 85.71267 |
| 19 | 9.63E+02 | 7.72E+00 | 9.06E+02 | 8.12E−01 | 9.16E+02 | 1.07E+00 | 2.77E+02 | 116.2018 | **2.00E+02** | 0.002584 |
| 20 | 9.60E+02 | 6.53E+00 | 9.06E+02 | 4.04E−01 | 9.16E+02 | 1.36E+00 | 2.07E+02 | 172.063 | **1.60E+02** | 81.33856 |
| 21 | 5.10E+02 | 3.45E+00 | 5.59E+02 | 1.79E+02 | 9.27E+02 | 1.37E+02 | **2.23E+02** | 182.8915 | 3.32E+02 | 170.8737 |
| 22 | 1.08E+03 | 2.19E+01 | 8.77E+02 | 1.04E+01 | 1.21E+03 | 1.99E+01 | 3.31E+02 | 120.8642 | **2.54E+02** | 55.26927 |
| 23 | 5.49E+02 | 2.56E+01 | 5.91E+02 | 1.72E+02 | 9.48E+02 | 1.38E+02 | **2.54E+02** | 223.3001 | 4.82E+02 | 140.5439 |
| 24 | 2.00E+02 | 3.48E−02 | 9.20E+02 | 1.70E+02 | 4.67E+02 | 6.23E+00 | 2.00E+02 | 5.65E−07 | **1.68E+02** | 43.36581 |
| 25 | 1.51E+03 | 8.75E+00 | 1.64E+03 | 3.33E+00 | 1.88E+03 | 4.44E+00 | 1.37E+02 | 5.65E−07 | **1.11E+02** | 12.35356 |

## 5.7 WOAGWO for Solving Pressure Vessel Design Problem

Pressure Vessel design is a classical engineering problem. The main goal of this problem is to optimize the cost of three sections of the cylindrical pressure vessel. Those sections should be minimized, which are forming, material and welding. The head of the vessel has hemi-spherical shape while the end of both sides of the vessel is crapped. This problem has four variables to optimize. These variables are shell thickness $T_s$, head thickness $T_h$, inner radius $R$, cylindrical length section without counting the head $L$. Therefore, this problem has four constraints that can be optimized. The following equations describe the constraints of the problem.

$$n = 1,2,3,4$$

$$\vec{x} = [x_1 x_2 x_3 x_4] = [T_s T_h\ R\ L],$$

$$f(\vec{x}) = 0.6224 x_1 x_3 x_4 + 1.7781 x_2 x_3^2 + 3.1661 x_1^2 x_4 + 19.84 x_1^2 x_3, \qquad (17)$$

Variable limitation

$$0 \leq x_1 \leq 99,$$

$$0 \leq x_2 \leq 99,$$

$$10 \leq x_3 \leq 200,$$

$$10 \leq x_4 \leq 200,$$

These are subjected to

$$g_1(\vec{x}) = -x_1 + 0.0193 x_3 \leq 0 \qquad (18)$$

$$g_2(\vec{x}) = -x_3 + 0.00954 x_3 \leq 0 \qquad (19)$$

$$g_3(\vec{x}) = -\pi x_3^2 x_4 - \frac{4}{3}\pi x_3^3 + 1{,}296{,}000 \leq 0 \qquad (20)$$

$$g_4(\vec{x}) = x_4 + 240 \leq 0 \qquad (21)$$

WOA achieved the best results for solving the problem [16]. Therefore, the authors used three metaheuristic algorithms to solve the problem, for example, WOA, WOAGWO, and FDO [54]. Table 12, shows that WOAGWO outperforms well compared to WOA and FDO. WOAGWO achieved results that are better than the other two algorithms. WOAGWO obtained these results 1.63, 1.43, 67.07, 10 for $T_s, T_h, R$ and $L$ respectively.

**Table 12 Comparison WOA, WOAGWO, and FDO for Pressure Vessel Design**

| WOA | | WOAGWO | | FDO | |
|---|---|---|---|---|---|
| **Avg.** | **Std.** | **Avg.** | **Std.** | **Avg.** | **Std.** |
| 1.36E+04 | 12671.54 | 1.32E+04 | 2536.893 | 5.33E+04 | 47583.22 |

# 6. Conclusion

To sum up, both WOA and GWO along with their modifications and hybridizations were presented. WOA and GWO with their limitations were highlighted. WOA and GWO with their algorithmic details were described in detail. The new approach "WOAGWO" was presented. The experimental results were explained to assess the performance of WOAGWO.

Several experiments were conducted to evaluate WOAGWO. WOAGWO was tested on 23 benchmark test functions to assess its performance in both exploitation and exploration. WOAGWO showed its superiority in 20 out of 23 functions compared to WOA and GWO. WOA and WOAGWO have the same result in 1 function, and WOA has slightly better than WOAGWO in 2 functions.

In addition, CEC2005 benchmark functions were used to evaluate WOAGWO. As a result, WOAGWO performed well in 14 functions. Though, it had the same result with WOA in 1 function. Nonetheless, WOA was better than WOAGWO in the other 3 functions. In spite of having a better overall result, WOAGWO was better than GWO in only 14 functions out of 25.

Furthermore, WOAGWO was evaluated by the CEC2019 benchmark function, and then results were compared to WOA and GWO. Consequently, WOAGWO had the same result with WOA in 1 function while it had better results in 7 functions. However, WOA performance was better than WOAGWO in 2 functions. WOAGWO was also compared to GWO, the results showed that WOAGWO was superior to GWO in 5 functions. They were also the same in 2 functions. In the face of having these results, GWO worked well in 3 functions.

Wilcoxon rank-sum test was used to evaluate the WOAGWO statistically, WOAGWO obtained significant results in 17 out of 23 benchmark functions. It was also tested on CEC2005 functions, so it achieved a better result in 13 functions. Furthermore, it had 6 significant results out of 10 by using CEC2019 test functions.

Then, WOAGWO was compared with DE, ABC, BSO, and WOA. Like WOAGWO, BSO was better in 9 benchmark functions. WOA was competitive in 4 functions. In addition, DE had a third rank in comparison. As a result, WOAGWO performance improved exploration capability. Overall, it can be said that WOAGWO improved the solution quality after each iteration, and it avoids local optima.

Finally, WOAGWO was used to solve a real world problem in the field of engineering. The problem was pressure vessel design which was solved by WOAGWO, WOA, and FDO. WOAGWO attained an optimum solution that was better than WOA and FDO.

Generally, WOAGWO improved the WOA standard and could improve solutions for those problems that were related to poor performance and dwindling into local optima in the exploration phase. WOAGWO produced significant results in almost all unimodal and multimodal functions. WOAGWO produced better results in the benchmark test functions because of the two techniques that were included in WOAGWO. Using the condition which was added inside the exploration phase to avoid whales to move to positions which were not better than the previous positions and also to improve the exploration performance. Embedding conditions, related to $a$ value and adapting four GWO equations in the exploitation phase of WOA, forced the whales to have better results. Improving the performance of WOAGWO over WOA also belonged to the exploitation ability of beta and delta wolves to save the best solutions and decreased $a$ value that tried to stop the movement of the prey in order to hunt it by the whales. Another reason behind this improvement was that a new condition was added to the exploration phase for updating the whale.

Finally, the following potential research work can be conducted in the future:

1) Solving real-world problems such as medical problems and other engineering problems.
2) Hybridizing different techniques to improve the current results.
3) Implementing chaotic maps on the proposed hybridization for further enhancement.

## Conflict of Interest
The authors declare that they have no conflict of interest.